\documentclass[letterpaper]{article} 
\usepackage{aaai23}  
\usepackage{times}  
\usepackage{helvet}  
\usepackage{courier}  
\usepackage[hyphens]{url}  
\usepackage{graphicx} 
\usepackage{natbib}  
\usepackage{caption} 
\usepackage{algorithm}
\usepackage{algorithmic}

\usepackage{newfloat}
\usepackage{listings}
\DeclareCaptionStyle{ruled}{labelfont=normalfont,labelsep=colon,strut=off} 
\floatstyle{ruled}
\newfloat{listing}{tb}{lst}{}
\floatname{listing}{Listing}
\usepackage{adjustbox}

\nocopyright 

\usepackage{amsfonts}
\title{Language-Informed Transfer Learning for Embodied Household Activities}
\author {
    Yuqian Jiang \textsuperscript{\rm 1}\thanks{Work completed during an internship with Amazon Alexa AI.},
    Qiaozi Gao \textsuperscript{\rm 2},
    Govind Thattai \textsuperscript{\rm 2},
    Gaurav Sukhatme \textsuperscript{\rm 2,\rm 3}
}
\affiliations {
    \textsuperscript{\rm 1} The University of Texas at Austin,  
    \textsuperscript{\rm 2} Amazon Alexa AI,
    \textsuperscript{\rm 3} University of Southern California\\
    jiangyuqian@utexas.edu, \{qzgao, thattg\}@amazon.com, gaurav@usc.edu
}

\begin{document}

\maketitle

\begin{abstract}

For service robots to become general-purpose in everyday household environments, they need not only a large library of primitive skills, but also the ability to quickly learn novel tasks specified by users. Fine-tuning neural networks on a variety of downstream tasks has been successful in many vision and language domains, but research is still limited on transfer learning between diverse long-horizon tasks. We propose that, compared to reinforcement learning for a new household activity from scratch, home robots can benefit from transferring the value and policy networks trained for similar tasks. We evaluate this idea in the BEHAVIOR simulation benchmark which includes a large number of household activities and a set of action primitives. For easy mapping between state spaces of different tasks, we provide a text-based representation and leverage language models to produce a common embedding space. The results show that the selection of similar source activities can be informed by the semantic similarity of state and goal descriptions with the target task. We further analyze the results and discuss ways to overcome the problem of catastrophic forgetting.

\end{abstract}

\section{Introduction}
Domestic service robots have been envisioned to help in a variety of household activities. Imagine a single robot that can be versatile enough from tidying up the rooms to playing with kids. Such a robot not only requires the sensing, navigation, and manipulation capabilities, but also needs to intelligently combine these skills to perform each activity as requested by the users.

Since every home is different, a simple library of pre-programmed tasks will hardly serve the purpose. For example, when a user wants to clean the kitchen cupboard, the specific goal conditions they would like to achieve will depend on their personal preferences and constraints of the environment. Does the robot re-arrange the dishes in a certain pattern? Does the robot dust the outside of the cupboard? The reality is that there could be an infinite number of combinations of goals, and a robot will most likely have to learn to solve new goals after it is deployed in the individual homes.

In this paper, we study the problem of learning novel user-specified household activities for a service robot that is shipped with pre-trained policies for a set of standard activities. We propose to learn the new activity by transferring from the policy of a similar activity. Our hypothesis is that the transfer can be more efficient than learning the new activity from scratch if their initial state and goal conditions are similar. Intuitively, a robot should be able to learn \textit{putting away cleaned dishes} efficiently if it has a good policy for \textit{cleaning kitchen cupboard}. Further, we can measure activity similarities by leveraging language models to embed their state and goal descriptions.

We test our hypothesis using the BEHAVIOR benchmark~\cite{srivastava2021BEHAVIORBenchmarkEverydaya}. BEHAVIOR simulates a large number of household activities for an embodied AI to learn. We first present a reinforcement learning (RL) approach to solve a subset of activities from scratch. The approach leverages text descriptions of the agent's current state and goal to allow the policies to operate in a common state space. We then initialize the learner with each of the pretrained policies when training it on a new activity, and evaluate the hypothesis that the transfer performance corresponds to the semantic similarity between the activity text descriptions. We present some initial results to show the potential of this approach for enabling versatile and adaptive home robots.

\section{Related Work}
Transfer learning leverages the knowledge learned in a source domain to improve the performance of a learner on the target domain. Transfer learning in reinforcement learning has been studied to transfer knowledge between different Markov Decision Processes (MDPs)~\cite{zhu2021TransferLearningDeep, taylor2009TransferLearningReinforcement}. While many approaches are evaluated in tasks with the same high-level goal and only different configurations in Mujoco, navigation, and Atari domains~\cite{barreto2017SuccessorFeaturesTransfer, schaulUniversalValueFunction}, a few recent transfer learning approaches have demonstrated positive transfer between distinct Atari games~\cite{rusu2016ProgressiveNeuralNetworksa, fernando2017PathnetEvolutionChannels}. Soemers et al. introduces an approach that transfers policy and value networks between distinct board games that have different action spaces~\cite{soemers2021TransferFullyConvolutional}. Encouraged by these successes, we propose to transfer RL policies among distinct embodied household activities which require high-level long-horizon reasoning about a large variety of goal conditions. Further, this work proposes to use language models on activity descriptions to inform the selection of source domains.

BEHAVIOR is a benchmark where embodied AI solutions are evaluated on household activities in a realistic physics simulation. The activities are selected from the American Time Use Survey to reflect the real distribution of household chores. There has been very little success using RL to solve BEHAVIOR in its original setting~\cite{srivastava2021BEHAVIORBenchmarkEverydaya}. In this paper, the method of providing the text-based, fully observable state representation is most similar to the work done by Shridhar et al. for the ALFRED benchmark~\cite{shridhar2021ALFWorldAligningText}.  

\section{Approach}

Our approach consists of two steps. In the first part, we introduce a text-based state representation for a RL agent to efficiently learn a set of diverse BEHAVIOR activities from scratch. The state representation is also in a common embedding space to allow easy knowledge transfer to other activities. In the second part, we introduce how these pre-trained policies are re-used for learning new activities, and test our hypothesis that the semantic similarity between activity descriptions can be used to predict transfer performances. 

\subsection{Learning Single Activities}

We introduce a different RL formulation from the original one in the BEHAVIOR benchmark, in order to speed up learning these activities using standard RL algorithms. 

\subsubsection{Text-Based State and Goal Representation}

Given the low RL performance in the original setting of BEHAVIOR, we take a similar approach to ALFWORLD~\cite{shridhar2021ALFWorldAligningText} by providing full observability of the logical state in the form of language. The simulator backbone of BEHAVIOR extracts logical predicates that describe the current states and relations of all objects in the world. We filter the logical predicates to the ones relevant to the activity, and use a template to generate text descriptions of the logical state. Similarly, the goal conditions are represented with text descriptions. Figure~\ref{fig:example_state} shows the initial state for one instance of the \textit{cleaning kitchen cupboard} activity. Figure~\ref{fig:example_goal} shows the goal definition of the \textit{cleaning kitchen cupboard} activity. There are two goals: 1) dust every cabinet and 2) move all cups to one cabinet and all bowls to the other. For the example initial state, there are two ways to ground the second goal based on how the cups and bowls are assigned to cabinets, and each grounding leads to a distinct set of subgoals.

\begin{figure}
\centering
\fbox{\begin{minipage}{\columnwidth}
{top\_cabinet\_47 is \textit{dusty}. top\_cabinet\_47 is \textit{next to} cup\_1. bottom\_cabinet\_41 is \textit{dusty}. bottom\_cabinet\_41 is \textit{on top} cup\_0. bottom\_cabinet\_41 is \textit{next to} cup\_0. bottom\_cabinet\_41 is \textit{next to} bowl\_1. countertop\_26 is \textit{under} bath\_towel\_0. countertop\_26 is \textit{in reach of} robot. countertop\_26 is \textit{in same room as} robot. bath\_towel\_0 is \textit{on top} countertop\_26. bath\_towel\_0 is \textit{in reach of} robot. soap\_0 is \textit{on top} countertop\_26. soap\_0 is \textit{in reach of} robot. bowl\_0 is \textit{on top} countertop\_26. bowl\_0 is \textit{in reach of} robot. bowl\_1 is \textit{inside} bottom\_cabinet\_41. bowl\_1 is \textit{next to} bottom\_cabinet\_41. cup\_0 is \textit{inside} bottom\_cabinet\_41. cup\_0 is \textit{next to} bottom\_cabinet\_41. cup\_1 is \textit{inside} top\_cabinet\_47. cup\_1 is \textit{next to} top\_cabinet\_47. room\_floor\_kitchen\_0 is \textit{in reach of} robot. room\_floor\_kitchen\_0 is \textit{in field of view of} robot.}

\end{minipage}}
\caption{An example initial state of \textit{cleaning kitchen cupboard}}
\label{fig:example_state}
\end{figure}

\begin{figure}
\centering
\fbox{\begin{minipage}{\columnwidth}
{For every cabinet, the following is NOT true: \\ the cabinet is dusty. \\
For at least one cabinet, for every bowl, the bowl is inside the cabinet,
and the following is NOT true: \\ cup1 is inside the cabinet. \\
For at least one cabinet, for every cup, the cup is inside the cabinet,
and the following is NOT true: \\ bowl1 is inside the cabinet.}

\end{minipage}}
\caption{An example goal definition of \textit{cleaning kitchen cupboard}}
\label{fig:example_goal}
\end{figure}

\subsubsection{Action Primitives}

The action space includes a set of discrete action primitives implemented in BEHAVIOR: \textsc{grasp}, \textsc{toggle on}, \textsc{toggle off}, \textsc{open}, \textsc{close}, \textsc{place inside}, \textsc{place on top}. Each action primitive takes a parameter that refers to an object. For example, \textsc{place inside}(cabinet\_0) means the robot will put the object currently in its gripper into the cabinet.

\subsubsection{Problem Formulation}
We formulate a BEHAVIOR activity as a Markov Decision Process denoted by the tuple  $\mathcal{M} = (\mathcal{S}, \mathcal{A}, \mathcal{P}, R)$. $\mathcal{S}$ is the space that consists of tokenized state and goal descriptions. $\mathcal{A}$ is the space of action primitives, parameterized by the objects relevant to the activity. $\mathcal{P}(\cdot|s,a)$ is the unknown stochastic transition probabilities. $R: \mathcal{S}\times\mathcal{A}\times\mathcal{S} \to \mathbb{R}$ is the reward function. Given the grounded subgoals of the activity, $R$ is defined as follows:
if $a$ is not executable at $s$, $R(s, a, s') = -1$; otherwise, let $g(s)$ be the number of subgoals satisfied in the state $s$, $R(s, a, s') = \frac{g(s')  - g(s)}{\textrm{total number of subgoals}} \cdot c$ where $c$ is a large constant. The reward function penalizes choosing action primitives that are not executable, such as \textsc{toggle off}(cup\_0), and generously rewards achieving new subgoals. 
The objective is to learn a policy $\pi: \mathcal{S} \to \mathcal{A}$ that maximizes the expected total reward. 

\subsubsection{Actor-Critic Policy}
\begin{figure}
    \centering
    \includegraphics[width=\columnwidth]{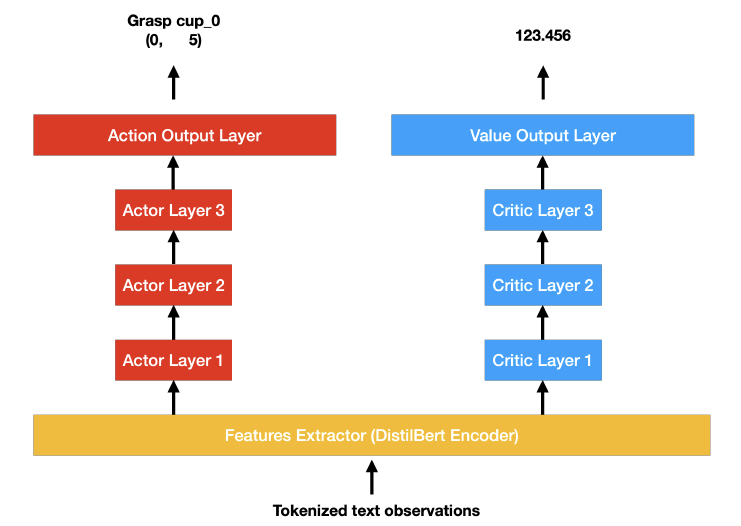}
    \caption{Actor-critic network architecture for learning one BEHAVIOR activity.}
    \label{fig:architecture}
\end{figure}

The policy can be trained by policy gradient methods such as PPO~\cite{schulman2017ProximalPolicyOptimization}.
Figure~\ref{fig:architecture} shows the actor-critic architecture. We use a pre-trained DistilBert model~\cite{sanh2020DistilBERTDistilledVersion} to tokenize and encode the input text. The actor network outputs a tuple of the action primitive index and the object index. 

\subsection{Transfer Learning}

Since the aim of this work is not to achieve top performances on BEHAVIOR, but rather to explore the connection between transfer performance and activity similarity, we adopt a straightforward method to re-use pre-trained policies and compare the learning curves.

\subsubsection{State and Action Mappings}

Since $S$ is a space of tokenized state and goal descriptions, the state space is common for all activities. However, the action primitives are parameterized by the objects in the scene, so the action space can have different sizes. To re-use a policy for a new activity, we copy all the weights in the network (Figure~\ref{fig:architecture}) except for the actor output layer. Then we resize the actor output layer to match the new action space and randomly initialize it before training.

\subsubsection{Semantic Similarity}

Given a new activity with an initial state and a set of goal conditions, the text-based state and goal representation constructed for the MDP formulation is also a unique description of this activity. We use the pre-trained SimCSE model~\cite{gao2022SimCSESimpleContrastive} to embed activity descriptions, and compute the consine similarity between the embeddings of any pair of activities.

\subsubsection{Transfer Metric}

We evaluate the transfer performance of each pair of activities by the transfer ratio (or transfer score) metric~\cite{taylor2009TransferLearningReinforcement,rusu2016ProgressiveNeuralNetworksa}. The transfer ratio measures the ratio of the total reward given to the transfer learner and the total reward given to the non-transfer learner after a certain number of training steps. It can be computed by the ratio of the area under the transfer learning curve over the area under the non-transfer learning curve.

\section{Experiments}

We choose to study 7 activities from BEHAVIOR: \textit{storing food}, \textit{cleaning kitchen cupboard}, \textit{putting away Halloween decorations}, \textit{collect misplaced items}, \textit{putting away cleaned dishes}, \textit{locking every window}, \textit{cleaning microwave oven}.


The policies are trained with the PPO algorithm as implemented in the stable-baselines3 library~\cite{stable-baselines3}. An episode terminates when all the subgoals are achieved or the maximum number of steps (64) has been taken. The hyperparameter $c$ in the reward function is set to 200. As a result, the highest total reward of an episode is 200, i.e. achieving all subgoals without any penalty. The lowest total reward is -64, i.e. always executing invalid actions.

\subsubsection{Training from Scratch}

To obtain a policy for each activity, we train for 512 episodes and take the top performing policy out of 3 runs.
Table~\ref{tab:scratch} shows the mean reward per episode achieved at the end of training by the top policy for each activity. Note that there is a wide gap between how well these activities are solved by our policies. The policies for \textit{locking every window} and \textit{cleaning microwave oven} are near optimal, whereas the policy for \textit{cleaning kitchen cupboard} never manages to achieve all subgoals during training. This difference is due to the solution length and the stochasticity of executing the action primitives. Some activities require executing more than 10 actions in the correct order, and some actions (e.g. grasp) have a low success rate in producing the desired effects. The uncertain action effects reflect the challenge for real robots, since the task-level policy should know how to recover when there are failures during execution.

\begin{table*}[t]
\centering
\begin{tabular}{||c c c c c c c||} 
 \hline
 food & cupboard & halloween & misplaced & dishes & window & microwave \\ [0.5ex] 
 \hline\hline
 -8.5 & -34.5 & 1.1 & 4.0 & -7.0 & 196.0 & 189.0 \\ 
 \hline
\end{tabular}
\caption{Mean reward per episode achieved at the end of training.}
\label{tab:scratch}
\end{table*}

Since it's much faster to learn \textit{window} and \textit{microwave} than the other activities, they are only used as source tasks but not target tasks in the transfer experiments below.

\subsubsection{Semantic Similarity}

\begin{figure}
    \centering
    \includegraphics[width=0.92\columnwidth]{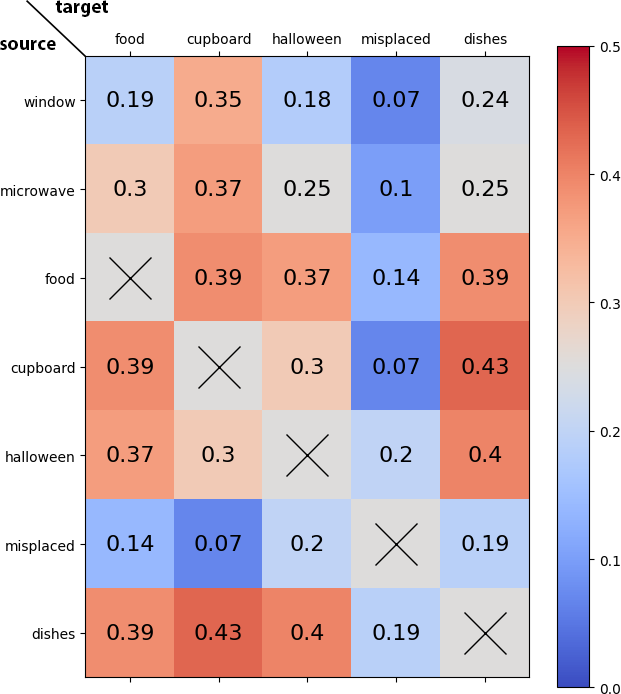}
    \caption{Semantic similarities between source and target activities.}
    \label{fig:semantic_similarity}
\end{figure}

Figure~\ref{fig:semantic_similarity} summarizes the semantic similarity in a matrix. 
Each row is a source activity and each column is a target activity. A high number (or warm color) means the descriptions of the two activities are close in the embedding space, whereas a low number (or cool color) indicates that the embeddings are distant. It may not be intuitive why some activities are more similar than others based on their abbreviated names. For example, \textit{storing food}, \textit{cleaning kitchen cupboard}, \textit{putting away dishes}, \textit{putting away Halloween decorations} all involve moving objects into cabinets, so their similarity scores are high when taking into account the full descriptions.

\subsubsection{Transfer Ratios}

\begin{figure}[t]
    \centering
    \includegraphics[width=0.92\columnwidth]{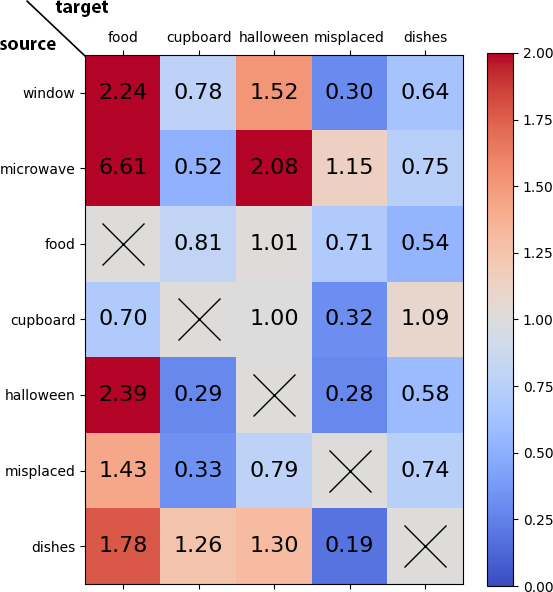}
    \caption{Transfer ratios of the first 80 episodes.}
    \label{fig:transfer_score_80}
\end{figure}

Figure~\ref{fig:transfer_score_80} presents the transfer ratio matrix after 80 episodes (or about 5000 steps). A ratio above 1 indicates positive transfer, i.e. the transfer learner receives higher total reward during training. Comparing with the similarity score matrix, we can make two observations. First, a high-quality source policy can lead to positive transfer, even if the activity is not similar. The activities \textit{storing food} and \textit{putting away Halloween decorations} (two difficult tasks) are not similar to \textit{locking every window} or \textit{cleaning microwave oven} (two easy tasks), but we see high transfer ratios in the first two rows of their columns. Second, for each target activity, higher semantic similarity has a higher chance of positive transfer. \textit{Cleaning kitchen cupboard} and \textit{putting away cleaned dishes} have a high semantic similarity (0.43). The only positive transfer to \textit{cupboard} was from \textit{dishes} and vice versa. On the other hand, \textit{collecting misplaced items} is semantically very different from all other activities, and gets some of the worst transfer ratios.

\subsubsection{Catastrophic Forgetting}

\begin{figure}[t]
    \centering
    \includegraphics[width=0.92\columnwidth]{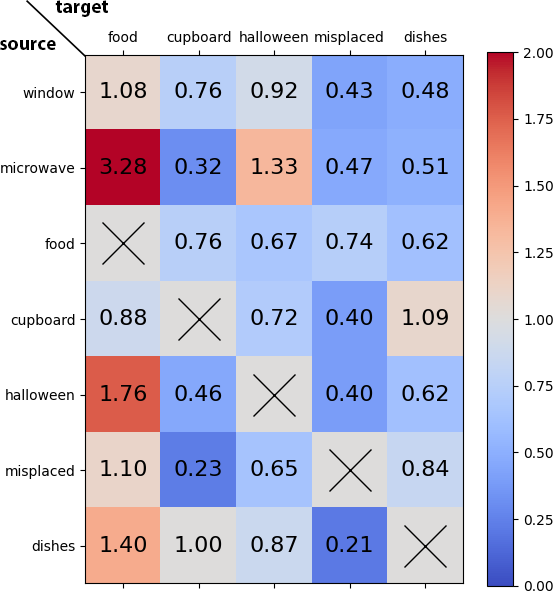}
    \caption{Transfer ratios of the first 160 episodes.}
    \label{fig:transfer_score_160}
\end{figure}

While there are clear signs that re-using policies can jump start learning a new activity, the benefits of transfer quickly disappear as catastrophic forgetting takes place. Figure~\ref{fig:transfer_score_160} shows the transfer ratios after 160 episodes (or about 10,000 steps). The general observations in Figure~\ref{fig:transfer_score_80} still hold, but the ratios are getting lower and there are fewer cases of positive transfer. 

For future studies, one of the ideas to transfer knowledge without suffering from the conflicting goals is by decoupling the task-independent knowledge from the task-dependent knowledge. 
In the case of household activities, there is a lot of shared knowledge across activities, especially the preconditions and effects of actions. For example, \textsc{toggle off}(cup\_0) is an invalid action in any activity. 
To this end, successor features~\cite{barreto2017SuccessorFeaturesTransfer} and universal value function approximation~\cite{schaulUniversalValueFunction} are both methods to learn representations that decouple the dynamics from the rewards so they will generalize over different goals.
Meanwhile, there are neural representations designed to avoid catastrophic forgetting. Progressive neural nets~\cite{rusu2016ProgressiveNeuralNetworksa} add a new column of network while preserving the weights learned in previous tasks.

\section{Conclusion}

We propose that home robots can efficiently learn novel household tasks from similar but distinct activities, and present our analysis in the BEHAVIOR benchmark. Our experiments show encouraging results: activity similarity measured by language embeddings can be used as a predictor for transfer performance, and a high-quality source policy of an easy but different activity can sometimes lead to a jump-start. We also observe the problem of catastrophic forgetting and suggest future research in this direction.

\bibliography{references}

\end{document}